\documentclass[sigconf]{acmart}
\usepackage{enumitem}
\usepackage{booktabs}
\usepackage{siunitx}
\usepackage{capt-of}
\usepackage[table]{xcolor}
\usepackage{graphicx}
\usepackage{subcaption}
\usepackage{graphicx}      
\usepackage{caption}
\usepackage[most]{tcolorbox}
\usepackage{tikz}
\usetikzlibrary{positioning}
\usepackage{url}
\usepackage{enumitem}
\setlist{noitemsep, topsep=0pt}
\definecolor{lavender}{HTML}{EEE8F6}
\definecolor{mint}{HTML}{E6F4EA} 
\definecolor{softRed}{HTML}{F8D7DA}
\definecolor{paperRed}{HTML}{FDE0E0} 

\usepackage{xspace}
\newcommand{\toolname}{TaskEval\xspace}

\title{\toolname: Synthesised Evaluation for Foundation-Model Tasks}

\author{%
  Dilani Widanapathiranage,
  Scott Barnett,
  Stefanus Kurniawan,
  Wannita Takerngsaksiri}
\email{{s224731539,scott.barnett,stefanus.kurniawan,wannita.takerngsaksiri}@deakin.edu.au}
\affiliation{%
  \institution{Applied Artificial Intelligence Initiative, Deakin University}
  \city{Geelong}
  \country{Australia}
}

\begin{document}

\begin{abstract}

Hallucinations are a key concern when creating applications that rely on 
Foundation models (FMs). Understanding where and how these subtle failures occur in an application relies on evaluation methods known as \textit{evals}. Prior work focuses on defining new eval methods or benchmark datasets for specific tasks. However, neither helps a software team with a task-specific FM application when there is no metric or dataset. The demand for both automated approaches and deep integration of human insight makes this a challenging problem. We address this gap by proposing an approach to synthesise a FM task-specific evaluator program that provides automation and a custom UI for capturing feedback. The core novelty of our approach lies in: (1) a task-agnostic meta-model that captures properties of any FM task, (2) an interaction protocol for efficient use of human feedback, and (3) an eval synthesiser that selects or generates an appropriate set of evals. We implement our approach in \toolname and demonstrate the concept on two diverse FM tasks: chart data extraction and document question answering. A preliminary evaluation on the quality of our selected evals shows 93\% and 90\% accuracy respectively. Our research tackles a growing problem facing engineering teams, how to evaluate and review outputs from FM tasks.

\end{abstract}

\maketitle

\section{Introduction}



\begin{figure}[t]
\vspace{2 mm}
  \centering
  
    \centering
    \includegraphics[width=\linewidth]{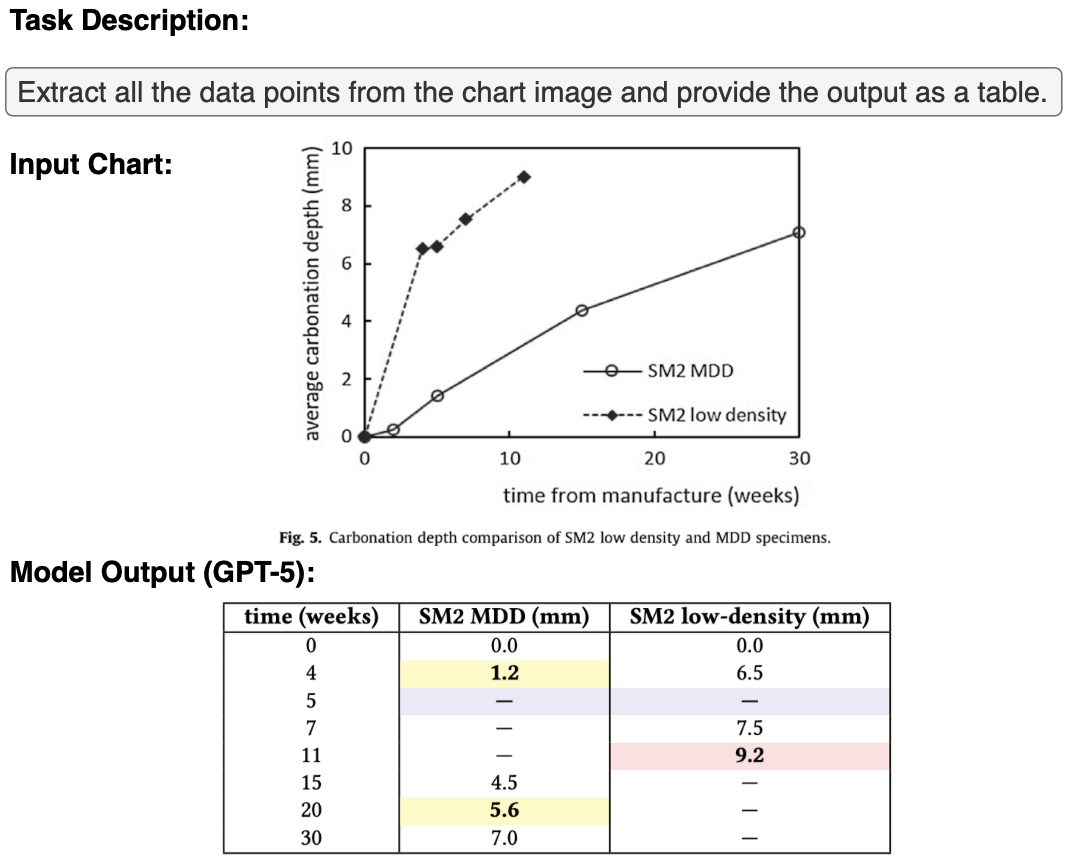}

 \caption{Chart-to-dataframe example~\cite{polak2025leveraging} with GPT-5. Human review required to identify: incorrect values (red), spurious values (yellow), and missing values (purple).}
 \label{fig:motivation-mini}
 \vspace{-5 mm}
\end{figure}

Complex tasks for software have become ubiquitous through advances in foundation models (FM). 
Consider Figure \ref{fig:motivation-mini} showing the task of converting an image of a chart to a table of records.
Prior to foundation models, this task requires a) extensive data collection, preparation, and labelling, b) computer vision and machine learning expertise, and c) continuous training and evaluation of machine learning models. With foundation models the same effect is achieved with a simple prompt in a few seconds. However, foundation models are prone to hallucinate causing subtle and unpredictable errors in the output (i.e. missing values and wrong values in Figure \ref{fig:motivation-mini}). 
A recent finding indicates that this can be reduced with better evals~\cite{kalai2025language} - metrics and processes to evaluate a foundation model on a task (an FM task). 
This problem now faces engineering teams who need domain and task-specific evals in order to build robust software. 

Evals have been extensively studied in the literature ~\cite{shankar2024validates,sriramanan2024llm,manakul2023selfcheckgpt,desmond2024evalullm,heo2025halucheck,li2025leveraging,wu2024large}. What is clear from this research is the importance of evals and the consideration of limitations of popular approaches such as LLM-as-a-judge -- where large language models (a type of foundation model) evaluate the output of foundation models ~\cite{wei2024systematic,thakur2024judging}. Most work on evals focuses on either presenting a new eval ~\cite{liu2023g,liang2022holistic} or creating a benchmark dataset~\cite{peng2025cweval,zheng2023judging,liang2024improving,white2024livebench} to assist with evaluating foundation models. Neither are helpful to software engineering teams who have a task-specific application for foundation models where there is no eval or available dataset. Existing work also does not consider the monitoring needs of continuous evaluation where a label is not available instantly~\cite{liang2022holistic,chen2025xverify}. The gap in the literature is there is no one technique that is a) task-agnostic, b) support no labelled examples, and c) supports operational requirements such as monitoring. We address this gap.



Our core idea is to synthesise an FM task-specific set of evals and UI for labelling examples. The novelty of our approach lies in 1) specifying a task-agnostic meta-model, 2) an interaction protocol for extracting task knowledge, and 3) an eval synthesiser that creates a custom eval or selects from a set of known evals. We have demonstrated the core elements of our approach in \toolname and conducted a preliminary evaluation of our approach on two diverse FM tasks (extracting data from a chart and question and answer use case). The output from \toolname includes an evaluator UI for labelling data and viewing eval results and an evaluator API for operational use cases such as CI. \toolname is designed to help teams bootstrap tooling for inspecting, labelling and evaluating their FM tasks. The research is guided by the following questions:





\begin{itemize}
    \item \textbf{RQ1:} How do we model FM tasks in a task-agnostic way?

    \item \textbf{RQ2:} How can we extract FM task details in a task-agnostic way? 
    
    \item \textbf{RQ3:} How can we synthesise task-specific evals for automatic evaluation and manual inspection? 
    
\end{itemize}

\section{Motivation}

Our research is motivated by a recent collaboration with a material scientist to use an FM to extract tabular data from charts. The following scenario highlights the core challenges. Consider Sarah, a software developer, tasked with creating a system to  extract data from charts for the purpose of analyzing crystalline structures and thermal properties. See Figure \ref{fig:motivation-mini} for an example. 

Sarah begins by using an FM for the task by writing a simple prompt to extract data from a chart. To evaluate the output Sarah needs a dataset with labels. As she is just starting, a dataset is not available so she needs to either create one herself with support of available tools (i.e. WebPlotDigitizer \cite{automeris}) or find a relevant opensource dataset (which is not always available). Sarah does not find a suitable dataset and has to create one herself. She also finds that existing evaluation frameworks are task-specific and don't capture the nuances of chart data extraction. 

After running the evaluation and comparing the extracted results against the labels Sarah finds the following issues: missing values, misidentified coordinates, and incorrect scales. Sarah is glad for her labelled dataset but realises that these errors are always going to occur. Sarah realises that even if her results are promising the material scientist will need to validate the output for correctness, full automation is only possible if the errors are acceptable.   

Sarah decides to create a tool to help the material scientist inspect the output from the chart data extraction task. This tool uses visualisations and a modified eval to help speed up the review process. This strikes a happy medium between full automation and human inspection. However, Sarah realises i) existing evals are task-specific, ii) evaluation strategies must be manually designed for each use case, and iii) each task requires custom tool and visualisations. 

To help Sarah, we propose \toolname, a framework that automatically synthesises a task-specific evaluator to support evaluation and human labelling of datasets.

\section{Vision}

\begin{figure*}[t]
  \centering
  \includegraphics[width=\textwidth]{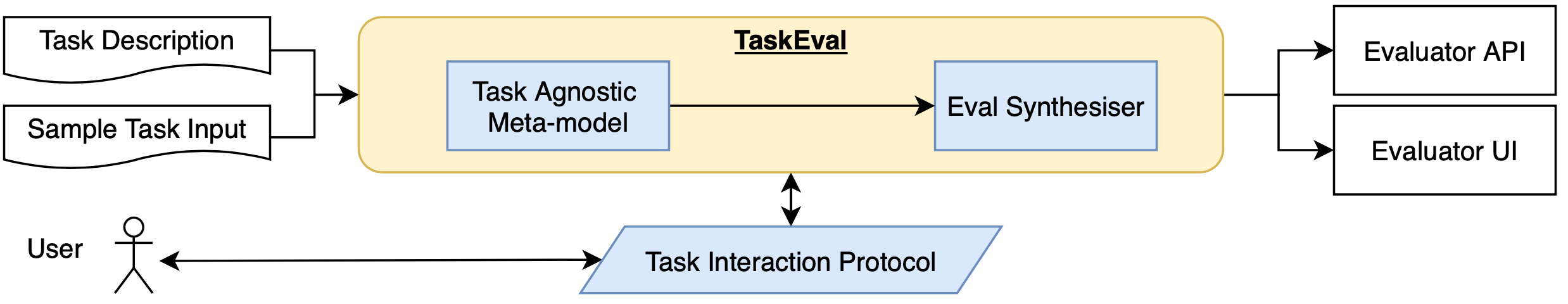}
  \caption{TASKEVAL system overview. Blue boxes show the novel elements of our solution: a) Task Interaction protocol, b) task-agnostic meta-model, and c) Eval Synthesiser. }
  \label{fig:genvalidator-overview}
  \vspace{2mm}
\end{figure*}

Our goal is to help engineering teams improve the quality of their evals of FM tasks. We achieve this by synthesising a task-specific evaluator and interface that supports human inspection. We implemented our ideas in \toolname to demonstrate the core elements. Our approach combines meta-modelling \cite{baudry2007model} from model-driven engineering (MDE)~\cite{bezivin2005model} with domain-modelling. An overview of the core elements of \toolname are shown in Figure~\ref{fig:genvalidator-overview}. Below, we describe the core components of our proposed solution. 





\vspace{-2mm}
\subsection{Task-agnostic Meta-model}. 
\label{sec:meta-model}

\noindent\fbox{%
  \parbox{\dimexpr\linewidth-2\fboxsep-2\fboxrule\relax}{%
    \textbf{RQ1. How do we model FM tasks in a task-agnostic way?}%
  }%
}

The \textbf{task-agnostic Meta-model} represents the essential properties of an FM task and elements required for creating an eval. \toolname accepts as input a description of the task and input for the task. Based on this information an instance of the task-agnostic meta-model is created and used to select appropriate evals. We anticipate needing to come up with criteria (and potentially new data fields) to adequately represent the context of an FM task. We plan to investigate multiple facets including: (i) input/output specification (modalities and formats), (ii) task type, (iii) evidence and grounding requirements, (iv) constraints (schema, units, ranges), (v) reasoning mode and answer multiplicity, (vi) evaluation objectives, and (vii) reference sources. The meta-model will define the relationships among these concepts. It is unlikely that a user will include this information on first pass, we plan to use an interaction protocol to iteratively populate the meta-model. 





\subsection{Task Interaction Protocol}
\label{sec:protocol}
\noindent\fbox{%
  \parbox{\dimexpr\linewidth-2\fboxsep-2\fboxrule\relax}{%
    \textbf{RQ2. How can we extract FM task details in a task-agnostic way? }%
  }%
}


The \textbf{Interaction Protocol} enables \toolname to incrementally instantiate the meta-model and to capture feedback. We present our protocol following guidelines from the literature~\cite{kurose2010computer}. It formalises how task knowledge is elicited, validated, and compiled into a runnable procedure, supporting:
(i) \emph{validation of potential errors}; 
(ii) \emph{validation or override of proposed strategies}; and 
(iii) \emph{refinement of evaluation requirements} via add/delete/edit operations.

\noindent\textbf{Message Types —} Request types include \textit{Validate Errors}, \textit{Update Evaluation Objective}, and \textit{Propose Strategies}; approval flow via \textit{Approve Plan}; execution via \textit{Run Evaluation}; and iteration via \textit{Provide Feedback}. Responses return structured feedback to \textit{approve}, \textit{reject}, or \textit{amend}.

\noindent\textbf{Syntax —} the syntax of our protocol is rooted in markdown that is both human and machine readable. Includes: \textit{task type}, \textit{io} (input/output), \textit{constraints}, \textit{objectives} (evaluation criteria), \textit{Potential errors}, and \textit{proposed strategies}.

\noindent\textbf{Field Semantics —} This includes specifies what the task output must align. \textit{Task type} selects evaluator templates; \textit{io} fixes modalities and formats; \textit{constraints} (schema/units/ranges) enable reference-free checks; \textit{objectives} specify what outputs must satisfy; \textit{potential errors} describe hypothesised failure modes; \textit{proposed strategies} bind descriptors to executable checks.

\noindent\textbf{Behavioural Rules —} The protocol follows a four-stage loop: Elicit (validate errors → feedback), Map (validate strategy mappings → feedback/objective update), Run (execute evaluation → results/error report), Refine (apply edits → update descriptor/plan). The loop repeats until the evaluation plan is finalised.

\subsection{Eval Synthesiser}
\label{sec:prog-synth}

\noindent\fbox{%
  \parbox{\dimexpr\linewidth-2\fboxsep-2\fboxrule\relax}{%
    \textbf{RQ3. How can we synthesise task-specific evals for automatic evaluation and manual inspection?}%
  }%
}

The \textbf{Eval Synthesiser} is responsible for synthesising and/or selecting an appropriate set of evals and determining the UI components required for displaying these evals. Choosing to create a custom eval or use an existing eval is based on the task and dimensions of evaluation that are expressed by the user. This approach enables our evals to include multiple dimensions. An important aspect of the UI generation is the consideration of how user feedback is captured. For example, in the table from chart example from Figure~\ref{fig:motivation-mini} a UI would need to be able to edit the extracted data when the FM makes an error. We design the Eval Synthesiser to ensure that labelled data is collected as people interact with the system.  Our demonstrator includes a set of eval strategies that are categorised as a) \texttt{summarize}, b) \texttt{visualize}, c) \texttt{Judge}, and d) \texttt{logic program} -- custom code generation. Each of these approaches includes pre-defined evals from the literature (i.e. rubric-based assessment such as G-Eval \cite{liu2023geval}). The Eval Synthesiser produces the Evaluator API and the Evaluator UI.  The API is designed for integration with operational infrastructure such as continuous integration pipelines.

\begin{figure*}[t]
  \centering
  
    \centering
    \includegraphics[width=\linewidth]{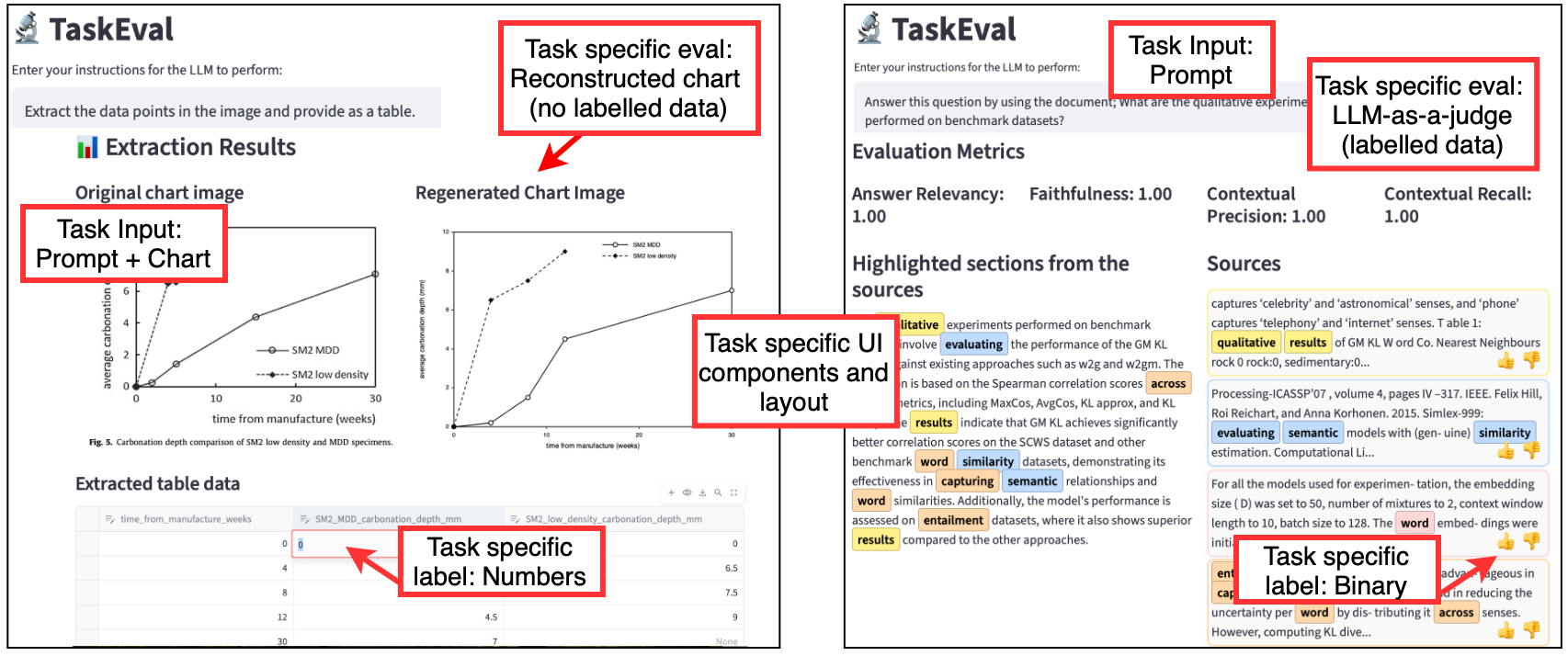}
    
    \label{fig:chart1}
 \caption{Output from \toolname for two tasks: chart data extraction (left) and document QA (right). Depending on task description, availability of labels and input data, \toolname selects task-specific a) evals (custom visualisation vs LLM-as-a-judge), b) human labels, and c) UI components and layout.}
\end{figure*}



\section{Why is it new? }


\textbf{Task-agnostic and modality independent.} Our approach works for any task including multi-dimensional tasks, objective or subjective, and FM or machine learning outputs. This is to support the next generation of FM Systems that use agents and techniques such as ML to solve problems. Unlike existing evaluation frameworks designed for specific domains or tasks, \toolname operates through a meta-model that captures task-agnostic descriptors, enabling synthesis of evaluations across diverse applications. This generalizability is essential as FM systems increasingly handle complex, multi-step problems where evaluation requirements vary significantly.

\textbf{Generate evaluation strategy and labelling UI} 


The oracle problem in FM evaluation, deciding correctness without ground truth remains unresolved. Existing approaches depend on costly human inspection ~\cite{feng2024sample,wang2025can} or on LLM-as-a-Judge methods that are inconsistent and themselves require evaluation. Prior tools are tied to benchmarks or labelled data, limiting reuse across tasks.
\toolname is new because it provides a ground-truth independent evaluation framework. It combines structured summaries, visualisations, automated judging, and evaluation logic in a unified UI. Unlike prior work, it supports continuous evaluation at inference time where labelled data is unavailable. Its reusable, project-agnostic design allows evaluation components to be adapted across domains without rebuilding logic.

\textbf{Evaluator API for use in development and application} \toolname generates an evaluation service API that is used by the interface to support engineering teams in evaluation. However, the API is also available for creating end user evaluation functionality enabling teams to focus on the application and not on the evaluation requirements. To the best of our knowledge this is the first approach that dual guides evaluation in development and with end user. 
\vspace{1 mm}
\section{\toolname Demonstration}

To demonstrate the framework, we implemented a proof-of-concept \toolname for two different tasks: (i) chart data extraction and (ii) question answering over documents. We conducted a preliminary evaluation of the ability for generating task-specific outputs and to assess the effort in evaluating the evals. Our tool synthesises task-specific evaluations based on the description of the task and a sample input. Figure~3 illustrates the generated UI for the two tasks and examples of evals.

\textbf{Task-specific generation of evaluations:}
\toolname uses a single pipeline with a shared meta-model. The meta-model captures essential elements of a task based on the task description and sample input. \toolname then selects relevant failure modes and binds them to reusable strategy templates. Figure~3 shows \toolname synthesising different evaluators for two FM tasks by adapting key attributes. \textbf{Task input:} \emph{Chart data extraction:} task description and a chart image, \emph{Document QA:} task description, source document, and expected answer(optional). \textbf{Task-specific eval:} \emph{Chart data extraction:} \textit{Visualise} strategy that redraws the chart from the extracted points for a side-by-side comparison with the original chart image. \emph{Document QA:} combines \textit{Visualise} and \textit{Judge} strategies, highlighting source spans that support answer claims and computing evaluation metrics (e.g., answer relevancy, faithfulness) via \texttt{DeepEval}\footnote{\url{https://github.com/confident-ai/deepeval}}. \textbf{Task-specific UI:} \emph{Chart data extraction:} side-by-side original and regenerated chart images with the extracted data table for inspection. \emph{Document QA:} displays highlighted evidence alongside the evaluation metrics. \textbf{task-specific label:} \emph{Chart data extraction:} numeric cell edits in the extracted table (e.g., correcting incorrect values from the motivating example in Figure~1). \emph{Document QA:} binary approvals on provided sources to confirm whether the answer is supported.

\textbf{Effort to evaluate the evals:}
\toolname generates code and other output using FMs which creates a problem of evaluating the evals. To evaluate the significance of this problem we performed a small experiment for each use case. For chart data extraction, we checked whether the regenerated plot matched the extracted points. On 30 charts, 28 were correct (93\%). For document QA, we verified whether the answer was mainly sourced from the passage with the highest count of highlighted words. Across 30 document QA tested, 27 met this criterion (90\%).










\vspace{-3.4mm}
\section{Future plans}
 
\textbf{Experimental evaluation of the generalisability of \toolname.} We plan to extend \toolname by supporting additional tasks including objective tasks (i.e. evaluation can be confirmed as right/wrong) and subjective tasks (i.e. multiple right answers). This will help us to discover the essential elements of a task and domain that needed to be modelled. We will use this information to update and refine our hierarchical task-agnostic meta-model (Section \ref{sec:meta-model}). Our evaluation plan is to compare our task-agnostic approach to other task \textit{specific} evaluation methods. This will help us discover the strengths and limitations of generalising across tasks. The outcome of this research will be an updated meta-model. 

\textbf{Evaluating the protocol with industry case studies.} 
The next stage of our research is to evaluate the protocol (Section \ref{sec:protocol}) through a series of industry case studies. We picked case studies as we want to understand how to improve the way practitioners evaluate FM tasks~\cite{runesonGuidelinesConductingReporting2009}. A case study will also enable us to gauge the preferences and benefits of our approach compared to manually creating evals and supporting tools. From collaborating with our industry partners we have found a wider range of FM tasks than are reported in the literature which will further provide insight into how we model the domain. The outcome from this evaluation will be refined protocol.   

\textbf{Optimise the creation of applications with FM tasks.}
The final stage of our research plan is to address a growing problem of \textit{how do we evaluate large amounts of generated code}~\cite{spiess2024calibration}. We will do this by 
extending our approach to optimise the end-to-end (i.e. design, implementation and testing) of any application with an FM task. \toolname will be extended with a) data synthesis capabilities based on prior work~\cite{sivasothy2024ragprobe}, b) include FM task generation and optimisation, and c) include a generate, test, and optimise loop. By focusing on evaluation first, our intention is to speed up the design and implementation of FM tasks providing a human guided system for application generation. The evaluation will focus on comparing human implemented applications with generated applications against multiple dimensions .

\vspace{-3.4mm}
\section{Risk} 

\textbf{Modelled task context does not generalise.} When creating \toolname we realised the challenge of capturing the context required for generating the evaluator programs. Without capturing sufficient context our approach will not generalise well to unseen FM tasks. To mitigate this risk our intention is to extend \toolname to many other tasks that include a) different input/output modalities, b) complexities, and c) types.  



\textbf{Variation of tasks is too vast to model.} For our approach to work we need to identify the essentials of multiple tasks to form a meta-model that enables the creation of tasks specific evaluation. This will require careful design of an agent that is able to work with engineering teams together to create evaluators. We believe this is an essential task moving forward as software increasingly writes the code the effort will shift to evaluation of the program. 

\textbf{Evaluating the evaluator.} Our approach uses FMs to create the evaluator which means the evaluator also needs to be evaluated! We plan to design a workflow and process that guides developers on how to create and evaluate the evaluator incrementally to minimise the effort required to evaluate the evaluator. 


\bibliographystyle{ACM-Reference-Format}
\bibliography{ref}

@article{polak2025leveraging,
  author  = {Maciej P. Polak and Dane Morgan},
  title   = {Leveraging Vision Capabilities of Multimodal {LLMs} for Automated Data Extraction from Plots},
  journal = {arXiv},
  year    = {2025},
  eprint  = {2503.12326},
  archivePrefix = {arXiv},
  url     = {https://arxiv.org/abs/2503.12326}
}

@article{feng2024sample,
  author={Feng, Kehua and Ding, Keyan and Tan, Hongzhi and Ma, Kede and Wang, Zhihua and Guo, Shuangquan and Cheng, Yuzhou and Sun, Ge and Zheng, Guozhou and Zhang, Qiang and others},
  title={Sample-efficient human evaluation of large language models via maximum discrepancy competition},
  journal={arXiv preprint arXiv:2404.08008},
  year={2024}
}

@article{wang2025can,
  author={Wang, Ruiqi and Guo, Jiyu and Gao, Cuiyun and Fan, Guodong and Chong, Chun Yong and Xia, Xin},
  title={Can llms replace human evaluators? an empirical study of llm-as-a-judge in software engineering},
  journal={Proceedings of the ACM on Software Engineering},
  volume={2},
  number={ISSTA},
  pages={1955--1977},
  year={2025},
  publisher={ACM New York, NY, USA}
}

@incollection{bezivin2005model,
  author={B{\'e}zivin, Jean},
  title={Model driven engineering: An emerging technical space},
  booktitle={International Summer School on Generative and Transformational Techniques in Software Engineering},
  pages={36--64},
  year={2005},
  publisher={Springer}
}

@misc{kurose2010computer,
  author={Kurose, James and Ross, Keith},
  title={Computer networks: A top down approach featuring the internet},
  year={2010},
  publisher={Pearson Addison Wesley}
}

@misc{automeris,
  author       = {{Automeris}},
  title        = {Automeris Digitization Tools},
  year         = {2025},
  howpublished = {\url{https://automeris.io/}},
  note         = {Accessed: 2025-09-18}
}

@article{runesonGuidelinesConductingReporting2009,
  title = {Guidelines for Conducting and Reporting Case Study Research in Software Engineering},
  author = {Runeson, Per and H{\"o}st, Martin},
  year = {2009},
  month = apr,
  journal = {Empirical Software Engineering},
  volume = {14},
  number = {2},
  pages = {131--164},
  issn = {1573-7616},
  doi = {10.1007/s10664-008-9102-8},
  urldate = {2025-01-13},
  langid = {english}
}

@article{sivasothy2024ragprobe,
  title={RAGProbe: An automated approach for evaluating RAG applications},
  author={Sivasothy, Shangeetha and Barnett, Scott and Kurniawan, Stefanus and Rasool, Zafaryab and Vasa, Rajesh},
  journal={arXiv preprint arXiv:2409.19019},
  year={2024}
}

@article{spiess2024calibration,
  title={Calibration and correctness of language models for code},
  author={Spiess, Claudio and Gros, David and Pai, Kunal Suresh and Pradel, Michael and Rabin, Md Rafiqul Islam and Alipour, Amin and Jha, Susmit and Devanbu, Prem and Ahmed, Toufique},
  journal={arXiv preprint arXiv:2402.02047},
  year={2024}
}

@inproceedings{liu2023geval,
  author    = {Liu, Tianyi and others},
  title     = {G-Eval: NLG Evaluation using GPT-4 with Better Human Alignment},
  booktitle = {EMNLP},
  year      = {2023},
  url       = {https://aclanthology.org/2023.emnlp-main.153.pdf}
}

@inproceedings{shankar2024validates,
  title={Who validates the validators? aligning llm-assisted evaluation of llm outputs with human preferences},
  author={Shankar, Shreya and Zamfirescu-Pereira, JD and Hartmann, Bj{\"o}rn and Parameswaran, Aditya and Arawjo, Ian},
  booktitle={Proceedings of the 37th Annual ACM Symposium on User Interface Software and Technology},
  pages={1--14},
  year={2024}
}

@article{sriramanan2024llm,
  title={Llm-check: Investigating detection of hallucinations in large language models},
  author={Sriramanan, Gaurang and Bharti, Siddhant and Sadasivan, Vinu Sankar and Saha, Shoumik and Kattakinda, Priyatham and Feizi, Soheil},
  journal={Advances in Neural Information Processing Systems},
  volume={37},
  pages={34188--34216},
  year={2024}
}

@article{manakul2023selfcheckgpt,
  title={Selfcheckgpt: Zero-resource black-box hallucination detection for generative large language models},
  author={Manakul, Potsawee and Liusie, Adian and Gales, Mark JF},
  journal={arXiv preprint arXiv:2303.08896},
  year={2023}
}

@inproceedings{desmond2024evalullm,
  title={EvaluLLM: LLM assisted evaluation of generative outputs},
  author={Desmond, Michael and Ashktorab, Zahra and Pan, Qian and Dugan, Casey and Johnson, James M},
  booktitle={Companion proceedings of the 29th international conference on intelligent user interfaces},
  pages={30--32},
  year={2024}
}

@article{heo2025halucheck,
  title={HaluCheck: Explainable and verifiable automation for detecting hallucinations in LLM responses},
  author={Heo, Sangwoo and Son, Sungwook and Park, Hyunwoo},
  journal={Expert Systems with Applications},
  volume={272},
  pages={126712},
  year={2025},
  publisher={Elsevier}
}

@article{li2025leveraging,
  title={Leveraging llms as meta-judges: A multi-agent framework for evaluating llm judgments},
  author={Li, Yuran and Mohamud, Jama Hussein and Sun, Chongren and Wu, Di and Boulet, Benoit},
  journal={arXiv preprint arXiv:2504.17087},
  year={2025}
}

@article{wu2024large,
  title={Large language models can self-correct with key condition verification},
  author={Wu, Zhenyu and Zeng, Qingkai and Zhang, Zhihan and Tan, Zhaoxuan and Shen, Chao and Jiang, Meng},
  journal={arXiv preprint arXiv:2405.14092},
  year={2024}
}

@inproceedings{peng2025cweval,
  title={Cweval: Outcome-driven evaluation on functionality and security of llm code generation},
  author={Peng, Jinjun and Cui, Leyi and Huang, Kele and Yang, Junfeng and Ray, Baishakhi},
  booktitle={2025 IEEE/ACM International Workshop on Large Language Models for Code (LLM4Code)},
  pages={33--40},
  year={2025},
  organization={IEEE}
}

@article{zheng2023judging,
  title={Judging llm-as-a-judge with mt-bench and chatbot arena},
  author={Zheng, Lianmin and Chiang, Wei-Lin and Sheng, Ying and Zhuang, Siyuan and Wu, Zhanghao and Zhuang, Yonghao and Lin, Zi and Li, Zhuohan and Li, Dacheng and Xing, Eric and others},
  journal={Advances in neural information processing systems},
  volume={36},
  pages={46595--46623},
  year={2023}
}

@article{liang2024improving,
  title={Improving llm reasoning through scaling inference computation with collaborative verification},
  author={Liang, Zhenwen and Liu, Ye and Niu, Tong and Zhang, Xiangliang and Zhou, Yingbo and Yavuz, Semih},
  journal={arXiv preprint arXiv:2410.05318},
  year={2024}
}

@article{white2024livebench,
  title={Livebench: A challenging, contamination-free llm benchmark},
  author={White, Colin and Dooley, Samuel and Roberts, Manley and Pal, Arka and Feuer, Ben and Jain, Siddhartha and Shwartz-Ziv, Ravid and Jain, Neel and Saifullah, Khalid and Naidu, Siddartha and others},
  journal={arXiv preprint arXiv:2406.19314},
  volume={4},
  year={2024}
}

@article{liu2023g,
  title={G-eval: NLG evaluation using gpt-4 with better human alignment},
  author={Liu, Yang and Iter, Dan and Xu, Yichong and Wang, Shuohang and Xu, Ruochen and Zhu, Chenguang},
  journal={arXiv preprint arXiv:2303.16634},
  year={2023}
}

@article{liang2022holistic,
  title={Holistic evaluation of language models},
  author={Liang, Percy and Bommasani, Rishi and Lee, Tony and Tsipras, Dimitris and Soylu, Dilara and Yasunaga, Michihiro and Zhang, Yian and Narayanan, Deepak and Wu, Yuhuai and Kumar, Ananya and others},
  journal={arXiv preprint arXiv:2211.09110},
  year={2022}
}

@article{wei2024systematic,
  title={Systematic evaluation of llm-as-a-judge in llm alignment tasks: Explainable metrics and diverse prompt templates},
  author={Wei, Hui and He, Shenghua and Xia, Tian and Liu, Fei and Wong, Andy and Lin, Jingyang and Han, Mei},
  journal={arXiv preprint arXiv:2408.13006},
  year={2024}
}

@article{thakur2024judging,
  title={Judging the judges: Evaluating alignment and vulnerabilities in llms-as-judges},
  author={Thakur, Aman Singh and Choudhary, Kartik and Ramayapally, Venkat Srinik and Vaidyanathan, Sankaran and Hupkes, Dieuwke},
  journal={arXiv preprint arXiv:2406.12624},
  year={2024}
}

@article{chen2025xverify,
  title={xverify: Efficient answer verifier for reasoning model evaluations},
  author={Chen, Ding and Yu, Qingchen and Wang, Pengyuan and Zhang, Wentao and Tang, Bo and Xiong, Feiyu and Li, Xinchi and Yang, Minchuan and Li, Zhiyu},
  journal={arXiv preprint arXiv:2504.10481},
  year={2025}
}

@inproceedings{baudry2007model,
  title={Model-driven engineering for requirements analysis},
  author={Baudry, Benoit and Nebut, Clementine and Le Traon, Yves},
  booktitle={11th IEEE international enterprise distributed object computing conference (EDOC 2007)},
  pages={459--459},
  year={2007},
  organization={IEEE}
}

@article{kalai2025language,
  title={Why language models hallucinate},
  author={Kalai, Adam Tauman and Nachum, Ofir and Vempala, Santosh S and Zhang, Edwin},
  journal={arXiv preprint arXiv:2509.04664},
  year={2025}
}

\end{document}